\documentclass[11pt]{article}
\usepackage[utf8]{inputenc}
\usepackage[T1]{fontenc}
\usepackage{eamt26}
\usepackage{times}
\usepackage{url}
\usepackage{latexsym}
\usepackage[small,bf]{caption}
\usepackage[table]{xcolor}
\usepackage{booktabs}
\usepackage{tabularx}
\usepackage{array}
\title{Artificial intelligence language technologies in multilingual healthcare: Grand challenges ahead}

\author{Vicent Briva-Iglesias\\
  School of Applied Languages and Intercultural Studies (SALIS)\\
  CTTS, ADAPT Centre\\
  Dublin City University\\
  {\tt vicent.brivaiglesias@dcu.ie}
}

\date{}

\begin{document}
\maketitle
\begin{abstract}
  AI language technologies (AILTs), increasingly enabled by large language models (LLMs), are becoming embedded in multilingual healthcare workflows for translation, rewriting, documentation, interpreting, and messaging in language-discordant settings. Yet fluent output is not the same as clinically safe or equitable communication: performance varies across languages, accents, tasks, and workflows, and efficiency gains can hide errors, reduce traceability, and shift responsibility across clinicians, translators, interpreters, and health systems. This narrative review synthesises recent peer-reviewed evidence across written communication, spoken communication, and emerging agentic workflows. Using the Human-Centered AI Language Technology (HCAILT) lens, it examines capabilities, evaluation practices, implementation patterns, and recurrent errors through reliability, safety culture, and trustworthiness. We identify key convergences and contradictions in the literature and propose seven grand challenges for the next phase of research and deployment. Progress, we argue, requires not only better models but also accountable sociotechnical design, calibrated human oversight, and stronger collaboration across MT/NLP, translation studies, HCI, clinical practice, implementation science, and policy.
\end{abstract}

\section{Introduction}
Multilingual healthcare is one of the clearest high-stakes environments in which AI language technologies can produce both meaningful benefit and meaningful harm. When clinicians and patients do not share a language, the consequences extend beyond inconvenience to comprehension, adherence, care continuity, and negative outcomes \cite{rawalAssociationLimitedEnglish2019,lionArtificialIntelligenceLanguage2024}. A recent systematic review and meta-analysis found that adult patients in language-discordant settings had higher odds of readmission and emergency department revisits, whereas access to verified interpretation attenuated those differences \cite{Chu2024}. Woods et al. \shortcite{Woods2022} likewise concluded that limited English proficiency is associated with poorer outcomes after hospital-based care, while van Lent et al. \shortcite{vanLent2025} found that shared language and professional interpreters continue to outperform informal interpreting and most digital tools in complex care situations, as supported by extensive literature \cite{Dew2018,genoveseArtificialIntelligenceClinical2024b,valdezMachineTranslationHospital2025a}.

At the same time, pressure to adopt AI-powered language technologies is increasing. Recent generative AI (genAI) systems can translate text, simplify complex documents, transcribe and summarise conversations, draft responses, and increasingly operate inside broader workflows linked to patient portals, documentation systems, and electronic health records \cite{brownLanguageModelsAre2020}. However, apparent fluency is an unreliable proxy for safety. A system may produce highly readable discharge instructions while mishandling dosage information, perform well in a dominant language while degrading sharply in a minor one, or reduce documentation time while introducing inaccuracies that are difficult to detect \cite{koeneckeCarelessWhisperSpeechtoText2024,leungAIScribesHealth2025}. For this reason, the central question is no longer whether these systems 'work' in a general sense, but for whom, in which language, for what task, under what workflow conditions, and with what consequences.

This paper addresses that question through a narrative review of recent peer-reviewed literature on AILTs in multilingual healthcare. A narrative review is appropriate because the evidence base is heterogeneous across modalities, user groups, tasks, settings, and outcomes, and because the field is evolving quickly enough that conceptual synthesis is as important as point-by-point benchmarking \cite{Sukhera2022}. The paper is intentionally user-centred in two senses. First, it focuses on the real users who encounter these systems: patients, clinicians, translators, interpreters, administrative staff, and healthcare organisations. Second, it examines the broader sociotechnical conditions that shape use, including interfaces, workflow design, escalation pathways, institutional accountability, and language inequity.

The paper has two linked aims. The first is descriptive: to synthesise the state of the art in multilingual healthcare AILTs across written communication, spoken communication, and emerging agentic workflows. The second is programmatic: to use that synthesis to articulate the major grand challenges that remain open for the next phase of research and deployment. In this sense, the paper is positioned not only as a healthcare AILTs review, but also as a contribution to the Translators and Users agenda in MT research: it asks how multilingual communication is being reconfigured by AI systems, what kinds of users are expected to rely on them, and what kinds of oversight, competencies, and institutional safeguards are required if these tools are to support rather than undermine safe care.

The review is selective rather than exhaustive. It focuses on recent peer-reviewed studies that examine AILTs used for multilingual healthcare communication across text, speech, and workflow orchestration, with priority given to work reporting empirical evaluation, implementation experience, or clinically relevant governance implications. The aim is not to catalogue every healthcare AILT paper involving language, but to identify the strongest current evidence, the main points of convergence and contradiction, and the unresolved problems that are most consequential for future research and deployment.

\section{Conceptual framing}
The grand-challenges tradition in HCI offers a useful way of structuring an interdisciplinary agenda for a field undergoing rapid technological change. Rather than listing isolated problems, grand-challenge papers identify cross-cutting tensions that span methods, stakeholders, and application domains. Stephanidis et al. \shortcite{Stephanidis2019} framed seven HCI grand challenges around issues such as accessibility, ethics, privacy, security, and health. Their second revisit argued that these challenges had not receded, but had instead been intensified by AI, particularly in relation to transparency, value alignment, explainability, user control, and accountability \cite{Stephanidis2025}. Their most recent update reaches a similar conclusion: genAI does not replace earlier human-centred concerns, but intensifies them around human autonomy, operational safety under non-deterministic outputs, accountability, governance-by-design, and alignment with human cognitive processes \cite{winslowRevisitingSixHumanCentered2026}. That framing is especially relevant to multilingual healthcare because questions of trust, human control, and explainability become concrete in language-discordant clinical encounters. They concern who understands a diagnosis, who detects a mistranslation, who reviews a rewritten discharge instruction, and who takes responsibility when an AI-mediated communication failure affects care.

A grand-challenge framing is useful here for three reasons. First, the literature on multilingual healthcare AILTs is fragmented across MT/NLP, translation studies, health communication, implementation science, and HCI. Second, the empirical evidence is uneven: some applications already have promising deployment studies \cite{tuConversationalDiagnosticArtificial2025a,gallifantFieldGuideDeploying2025}, while others remain early, speculative, or weakly evaluated in naturalistic tasks \cite{mellingerChapter2Designing2025}. Third, healthcare adoption depends on much more than model quality. It also depends on workflow design, training, monitoring, reporting, procurement, and institutional legitimacy. A grand-challenge framing helps keep those dimensions visible at the same time.

If grand challenges provide the paper's structure, the Human-Centered AI Language Technology (HCAILT) framework provides its analytical lens \cite{BrivaIglesias2026}. HCAILT adapts broader human-centred AI thinking to multilingual language technologies and foregrounds three linked pillars: reliability, safety culture, and trustworthiness. Reliability refers to whether a system performs consistently across languages, tasks, and conditions, and whether it is fit for a particular communicative purpose. Safety culture refers to the organisational practices that anticipate failure rather than assuming success, including monitoring, auditing, literacy, incident reporting, role clarity, and explicit escalation. Trustworthiness refers not to how much users happen to trust a system, but to whether that trust is warranted because the system's behaviour can be inspected, governed, challenged, and corrected. Recent work on medical AI trustworthiness reinforces this distinction, arguing that trustworthy systems require not only technical performance but institutional infrastructures for oversight and contestability \cite{Goisauf2025,Shneiderman2020,vanLeersum2025}.

In this paper, AILTs in multilingual healthcare are defined broadly as "AI systems that process or generate language across text, speech, or multimodal inputs and outputs in ways that affect multilingual healthcare communication". The emphasis falls on three practical domains: written communication, spoken communication, and agentic workflows. The central concern is therefore not simply what has been built, but what kinds of communicative problems remain unsolved when these systems are examined through the combined lenses of UX, clinical risk, and multilingual inequity.
\section{State of the art in AI language technologies for multilingual healthcare}
This section reviews the current state of the art across three domains of use: written communication, spoken communication, and agentic workflows. Across all three domains, the literature points to the same broad pattern. Technical capability has improved rapidly, but performance remains highly task- and language-dependent, and the most consequential questions now concern workflow design, evaluation, governance, and human oversight rather than generation quality alone.

\subsection{Written communication}
Written communication remains the most mature and empirically developed area of AILT use in multilingual healthcare \cite{Dew2018,genoveseArtificialIntelligenceClinical2024b}. This is partly because text-based tasks map onto existing organisational bottlenecks: discharge instructions must be delivered quickly, educational materials need broad language coverage, and document workflows are easier to evaluate than live spoken encounters. At the same time, written communication is not a single task. It includes direct translation of clinical, specialised documents, translation of generic patient education and public-health materials, patient-facing information that needs to be understood, and hybrid workflows that combine automation with expert review.

\subsubsection{Machine translation of patient-specific clinical documents}
Earlier work already suggested that MT could support healthcare communication, while also stressing the limitations of general-purpose systems. Dew et al. \shortcite{Dew2018} identified promise for MT in health communication but called for stronger domain adaptation and more meaningful evaluation. Zeng-Treitler et al. \shortcite{ZengTreitler2010} similarly showed that access to translated medical content is not enough on its own if the resulting text remains difficult to understand. Herrera-Espejel and Rach \shortcite{HerreraEspejel2023}, reviewing public-health and epidemiological communication, argued that MT is becoming increasingly useful for outreach, but only when its limitations are understood and human validation remains central in higher-risk contexts.

Recent work on patient-specific discharge materials is more ambitious and more mixed. Ray et al. \shortcite{Ray2025} evaluated GPT-4o translations of personalised paediatric patient instructions into Spanish and found performance comparable to professional translation under an MQM-style framework. That is an important result, because it suggests that for a high-resource language and a constrained document type, contemporary LLM-based translation can approach professional-quality output. However, other studies complicate any simple parity narrative. Martos et al. \shortcite{Martos2025} compared AI-generated and professionally translated discharge instructions across Spanish, Chinese, Vietnamese, and Somali, and found that AI was non-inferior only for Spanish adequacy and error severity. Brewster et al. \shortcite{Brewster2025} likewise reported substantial variation across Arabic, Armenian, Bengali, Chinese, Somali, and Spanish, with weaker performance in digitally underrepresented languages.

Taken together, these studies suggest that the relevant question is not whether LLM translation works in healthcare in the abstract, but in which language pairs, for which document types, under what review conditions, and with what acceptable level of risk \cite{pymRiskManagementTranslation2025}. They also point to an important methodological shift: evaluation is moving away from generic fluency scoring and toward clinically meaningful error categorisation. That shift matters because a minor stylistic awkwardness and a dosage omission do not have the same consequences. In multilingual healthcare, surface quality is therefore an insufficient basis for deployment decisions.

\subsubsection{Post-editing translation workflows}
A second major lesson from the written literature is that the most useful comparison is often not AI versus humans, but one workflow configuration versus another. Brewster et al. \shortcite{Brewster2025} showed that what they call "human-in-the-loop" (hereafter, post-editing or "PE", since we think the HITL concept dehumanises the user) strategies can produce translations comparable to, or better than, professional translation alone while being substantially faster than a fully manual process. This is nothing new within the MT and Translation Studies communities (see, for example, Terribile \shortcite{terribilePosteditingReallyFaster2024}), but may have not been known in the medical informatics field. These findings are especially relevant for healthcare implementation because they suggest that the central design question is not whether AI should replace professional language workers, but how automation should be positioned within a reviewed and accountable pipeline, and that the crucial stakeholders - MT/TS communities and medical informatics - are not aware of what the others are doing.

Lopez et al. \shortcite{Lopez2025} make a related argument from an implementation perspective. Rather than focusing only on benchmark quality, they emphasise the organisational work required for safe deployment: integration with documentation and translation workflows, terminology control, auditability, clear lines of responsibility, and evaluation of patient comprehension rather than translation quality alone. This helps reconcile apparent contradictions across the literature. Studies such as Ray et al. \shortcite{Ray2025} show that raw model performance may already be strong enough to reduce manual effort in some high-resource settings. Studies such as Martos et al. \shortcite{Martos2025} and Brewster et al. \shortcite{Brewster2025}, however, show that residual risk remains substantial across the broader multilingual spectrum. Once differences in language set, task, and workflow are taken seriously, the broader conclusion becomes clearer: AI can already support safer and faster multilingual document production, but not yet in ways that justify removing expert oversight. 

\subsubsection{Plain language and rewriting}
Plain-language transformation and rewriting are central to multilingual accessibility because accurate translation alone may still leave patients with text that is technically correct but difficult to understand \cite{hansen-schirraEasyLanguagePlain2020a}. Recent evidence suggests that generative AI can be useful here, although again under bounded conditions. Zaretsky et al. \shortcite{Zaretsky2024} found that genAI could substantially improve readability and comprehensibility of discharge content, while also emphasising the need for better accuracy, completeness, and clinician review. Briva-Iglesias and Peñuelas-Gil \shortcite{briva-iglesiasSimplifyingHealthcareCommunication2025a} also shared similar results in informed consent forms via automatic readability metrics, and Rust et al. \shortcite{Rust2025} reported similar improvements in cardiology discharge-summary simplification, though with limitations in personalisation.

Findings are also encouraging, though conditional, for broader informational materials. Ugas et al. \shortcite{Ugas2025} found that human translation still outperformed MT across languages, even when MT often produced acceptable quality. Chen et al. \shortcite{Chen2025a}, evaluating critical-care educational content in Mandarin, Spanish, and Ukrainian, also showed meaningful access gains alongside substantial platform- and language-dependent variation. McMinn et al. \shortcite{McMinn2025} reported that a bespoke AI process could produce more readable first drafts of scientific plain-language summaries than medical-writer workflows alone.

These studies show that rewriting is a central component of equitable health communication. They also blur the boundary between translation and authorship, because these systems do not merely transfer content across languages, they also reshape it for particular users and contexts \cite{montalt-resurreccioPatientCentredTranslationCommunication2024}. In practice, this means that translation and plain-language rewriting should be treated as linked communicative tasks, both of which require context-aware review when clinical nuance or risk is involved.

\subsubsection{Main learnings and pitfalls in written communication}
Across the written literature, five lessons stand out. First, performance can already be very strong in high-resource languages and relatively constrained document types. Second, quality remains uneven across languages, especially for digitally underrepresented ones. Third, the strongest implementation evidence increasingly favours reviewed PE workflows over either raw automation or fully manual extremes. Fourth, readability and translation quality must be treated as distinct outcomes. Fifth, many of the hardest problems are organisational rather than purely technical, including terminology governance, workflow integration, quality assurance, and role allocation. This is where HCI methods should come into play.

The main pitfalls are equally consistent: clinically meaningful omissions, mistranslations hidden by fluent output, and evaluations that focus on text quality while neglecting patient comprehension and correct action. There is also a clear equity risk. If AI translation transforms service delivery in dominant languages while remaining unreliable in minor ones, multilingual healthcare may become more efficient and more unequal at the same time.

\subsection{Spoken communication}
If written communication is currently the most mature domain of AILT deployment, spoken communication is the most revealing one. Speech exposes the complexity of real-world multilingual healthcare: accent variation, code-switching, overlapping talk, noise, incomplete utterances, and emotionally charged exchanges. For that reason, the spoken literature is especially valuable for understanding end-to-end system risk.

\subsubsection{Machine interpreting}
Professional interpreting remains the benchmark for complex spoken multilingual care. van Lent et al. \shortcite{vanLent2025} make this clear: shared language and professional interpreters generally outperform informal interpreters and most digital tools in situations involving complexity, nuance, or clinical risk. That does not mean digital tools are irrelevant, but it does suggest that their role is currently best understood as bounded and task-contingent rather than universally substitutive. 

Recent deployment studies support this interpretation. Olsavszky et al. \shortcite{Olsavszky2025} piloted a digital translation platform designed to support consultations via multilingual mediation and found the platform feasible, but also identified personnel availability as a major bottleneck. Kothari et al. \shortcite{Kothari2025}, by contrast, described a system-wide digital medical interpretation framework focused on infrastructure, EHR integration, hardware, and operational scale. Together, these studies show that multilingual spoken communication is not only a model-performance problem. It is also a systems-design, procurement, and workflow problem.

The literature on direct speech translation in healthcare remains thinner and methodologically harder to interpret than the literature on written translation. Today, most research on spoken AILTs has focused on computer-assisted interpreting (CAI) and not machine interpreting \cite{luMachineComputerassistedInterpreting2025}. Even so, Iranzo-Sanchez et al. \shortcite{IranzoSanchez2025} showed in multilingual medical education that domain-adapted ASR and speech translation pipelines can substantially outperform general systems. This is encouraging evidence for the value of domain adaptation, but it still comes from settings that are more controlled than typical bedside care. Spoken multilingual healthcare therefore remains an area where technical progress is real, but deployment claims should remain cautious.

\subsubsection{Ambient scribes}
Ambient scribes are currently the most visible application of spoken healthcare AILTs. They extend the earlier digital scribe concept \cite{Coiera2018} by combining ASR, speaker diarisation, and LLM-based note generation. Recent studies suggest clear benefits, but also show why these systems should not be evaluated on perceived usefulness alone.

Balloch et al. \shortcite{Balloch2024} found that an ambient AI documentation tool improved documentation quality scores, shortened consultations, and reduced task load in simulated encounters. Stults et al. \shortcite{Stults2025} reported improved clinician satisfaction and reduced time spent on note-writing after deployment, while Olson et al. \shortcite{Olson2025} found reductions in burnout, cognitive load, and after-hours documentation. Shah et al. \shortcite{Shah2025} likewise reported positive clinician perceptions regarding workload and patient engagement.

However, the picture is less reassuring when the outcome shifts from experience to fidelity. Lukac et al. \shortcite{Lukac2025} found only modest reductions in documentation time and reported persistent accuracy concerns, including occasional clinically significant inaccuracies. Wang et al. \shortcite{Wang2025} proposed a formal evaluation framework showing that fluent notes can coexist with weaknesses in transcription, diarisation, factual accuracy, and medication capture. These findings are especially important for multilingual healthcare because each additional processing step introduces another opportunity for error.

From a multilingual perspective, ambient scribes raise a further concern. Although many are evaluated in predominantly monolingual documentation settings, their architecture is readily repurposed for multilingual encounters. Once that happens, the system may be transcribing accented English, patient speech in another language, interpreted speech, or machine-translated speech. The resulting error stack is cumulative. The literature therefore supports a cautious conclusion: ambient tools may already reduce administrative burden, but there is much weaker evidence that they are ready for multilingual clinical communication.

\subsubsection{Main learnings and pitfalls in spoken communication}
Spoken-language AI is now infrastructural in healthcare, because ASR underpins dictation, ambient documentation, conversational agents, and speech-to-speech systems. Across the literature, three points are clear. First, speech tools can improve usability and reduce burden, especially for documentation-related tasks. Second, domain adaptation matters. Third, performance remains highly context-dependent, and human review is still necessary in high-stakes settings \cite{Ng2025}.

The major risks are both technical and equity-related. Koenecke et al. \shortcite{Koenecke2020} showed racial disparities in ASR outside healthcare, while Zolnoori et al. \shortcite{Zolnoori2024} reported analogous disparities in patient-nurse communication. For multilingual care, this expands the equity problem beyond named languages to include accent, dialect, conversational style, and racialised speech. The strongest implication is methodological: spoken systems should be evaluated as end-to-end pipelines rather than as isolated components, because compounding errors across recognition, translation, summarisation, and note generation can distort the clinical record even when workflow efficiency appears to improve. This is something that most reviewed papers currently lack.

\subsection{Agentic workflows}
The third domain, agentic workflows, is empirically the least mature but strategically the most consequential. Here the focus shifts from single-task systems to orchestrated pipelines that combine language processing with retrieval, planning, routing, documentation, and interaction with clinical systems \cite{briva-iglesiasAreAIAgents2025}. In multilingual healthcare, this may include tools that receive patient messages, detect language, translate content, retrieve relevant context, draft a response, and route the case onward, or systems that process spoken encounters into notes and structured records \cite{heydariAnatomyPersonalHealth2025}.

This area matters because multilingual care is rarely experienced as a series of isolated language tasks. Patients need symptom intake, appointment preparation, portal communication, after-visit summaries, navigation guidance, and follow-up messaging. Clinicians need support with note drafting, inbox management, referral text, and multilingual communication. Agentic workflows promise to connect these tasks, but they also risk blurring task boundaries and obscuring where accountability lies \cite{ojewaleAIAccountabilityInfrastructure2025}.

\subsubsection{Evidence base and current maturity}
The empirical literature on real-world clinical LLM workflows remains relatively thin. Artsi et al. \shortcite{Artsi2025}, in a systematic review of real-world clinical workflows, found surprisingly few peer-reviewed empirical studies despite the high level of public attention surrounding LLM and AI agents deployment. Reported applications included message drafting, outpatient communication, mental health support, and information extraction, and some studies reported gains in efficiency and user satisfaction. At the same time, the review highlighted limited generalisability, regulatory delays, and a lack of robust post-deployment monitoring.

Chen et al. \shortcite{Chen2025b} similarly argue that LLMs and agents in healthcare require richer evaluation frameworks than traditional task-based systems. In multilingual healthcare this matters especially because, as previously discussed, a workflow may combine translation, retrieval, summarisation, and action. A system may perform reasonably at each subtask in isolation while still failing as a workflow if it loses negation during translation, retrieves outdated policy information, or generates an overconfident patient-facing summary.

The current evidence therefore supports a cautious position. Agentic systems may offer operational value, but the literature is presently stronger on potential than on mature multilingual clinical deployment. For this reason, the most defensible stance is neither dismissal nor exuberance, but tightly governed experimentation.
\subsubsection{Why agentic workflows are especially important for multilingual healthcare}
Multilingual healthcare is especially likely to benefit from agentic designs because language-discordant care often requires chains of actions rather than isolated outputs \cite{heydariAnatomyPersonalHealth2025}. A patient portal message, for example, may require language identification, translation, urgency recognition, retrieval of relevant medication information, drafting of a plain-language response, and routing to the appropriate team. Similar logic applies to discharge preparation, appointment reminders, consent support, and navigation tasks.

At the same time, multilingual healthcare is one of the least forgiving environments in which to assume that agentic flexibility is automatically beneficial. Language tasks may sit close to legal, ethical, and clinical thresholds. A mistranslated symptom, an over-smoothed explanation, or an inferred but undocumented detail can produce downstream harm. Agentic systems also raise the risk of silent task expansion: a tool perceived as a translation assistant may also begin summarising, simplifying, or prioritising without the user fully noticing that the communicative task has changed.

\begin{table*}[t]
\centering
\footnotesize
\setlength{\tabcolsep}{5pt}
\renewcommand{\arraystretch}{1.18}
\rowcolors{2}{black!3}{white}
\begin{tabularx}{\textwidth}{>{\raggedright\arraybackslash}p{0.26\textwidth} >{\raggedright\arraybackslash}X >{\raggedright\arraybackslash}p{0.20\textwidth} >{\raggedright\arraybackslash}p{0.16\textwidth}}
\toprule
\rowcolor{black!12}
\textbf{Grand challenge} & \textbf{What is at stake} & \textbf{Primary modalities} & \textbf{HCAILT linkage} \\
\midrule
\textbf{1. Clinically valid, risk-sensitive evaluation} & Preventing fluent-but-harmful output; demonstrating comprehension and correct action. & Text, speech-to-text, speech-to-speech & Reliability + Trustworthiness \\
\textbf{2. End-to-end multilingual fidelity} & Preventing cumulative errors across recognition, translation, rewriting, and summarisation. & Speech, text, multimodal pipelines & Reliability \\
\textbf{3. Bounded agency and safe failure} & Ensuring that flexible systems stay within scope and escalate appropriately. & Agentic workflows & Safety culture \\
\textbf{4. Redesign of human roles and competencies} & Defining who uses, reviews, governs, and takes responsibility for AI-mediated language work. & Text, speech, workflows & Safety culture + Trustworthiness \\
\textbf{5. Equity for minor languages, dialects, and accents} & Preventing a two-tier communication infrastructure. & All & Reliability + Safety culture \\
\textbf{6. Governance, regulation, and reporting} & Creating institutional and regulatory mechanisms for non-deterministic systems. & All & Safety culture \\
\textbf{7. Trust-oriented UX and overreliance prevention} & Designing interfaces that support calibrated use rather than blind acceptance. & All & Trustworthiness \\
\bottomrule
\end{tabularx}
\caption{Proposed grand challenges for AI language technologies in multilingual healthcare}
\end{table*}

The HCAILT lens is useful here because it brings these risks into a single frame \cite{BrivaIglesias2026}. Reliability requires that each step in the chain be fit for purpose and that end-to-end behaviour be evaluated rather than inferred from component quality. Safety culture requires explicit scope boundaries and escalation pathways. Trustworthiness requires that users understand what the system has done, what sources it has used, and where uncertainty remains.

\subsubsection{Governance, non-determinism, and bounded agency}
The governance literature suggests that agentic clinical systems may require new regulatory categories and stronger operational safeguards. Tan et al. \shortcite{Tan2026} argue that general-purpose, non-deterministic, increasingly agentic software fits poorly within traditional medical-device paradigms. This concern is reinforced by Winslow et al. \shortcite{winslowRevisitingSixHumanCentered2026}, and is especially relevant to multilingual language agents, which are often flexible, prompt-sensitive, and repurposable rather than narrowly locked systems.

In practical terms, this means that a multilingual agent linked to a patient portal or EHR should not treat translation, simplification, summarisation, triage framing, and explanation as interchangeable tasks. These activities have different risk profiles and should be bounded accordingly. See, for example, the EU AI Act's tier risk \cite{europeanunionEuropeanUnionAI2024}. Useful safeguards may include constrained retrieval, task-specific thresholds, confidence-aware escalation, editable intermediate outputs, provenance displays, and explicit human fallback. The critical question is not whether "humans remain symbolically in the loop", according to some, but whether they are genuinely positioned to detect and correct consequential failure.

\subsubsection{Main learnings and pitfalls in agentic workflows}
The main lesson from the current agentic literature is that the field is advancing conceptually faster than empirically. The strongest evidence still comes less from mature multilingual patient-care deployments than from workflow reviews, governance papers, and early implementation studies in adjacent clinical uses \cite{liuDeploymentcentricMultimodalAI2025,karunanayakeNextgenerationAgenticAI2025}. This does not reduce the importance of the area. It suggests that now is the right time to shape expectations before unsafe assumptions solidify.

The main pitfalls are opaque task boundaries, weak post-deployment monitoring, and a tendency to overinterpret workflow automation as a purely technical gain. In multilingual healthcare, agentic systems are attractive because they can connect communicative tasks that currently fragment care, but their danger lies in connecting those tasks without sufficient transparency, constraint, and accountability.

\section{Grand challenges ahead}
The literature reviewed does not point to a single bottleneck. It reveals a cluster of interlocking tensions that cannot be solved by model improvement alone. These are best understood as grand challenges because they span modalities, disciplines, and institutional layers. Table 1 summarises the seven challenges proposed here.

\subsection{Clinically valid, risk-sensitive evaluation}
The first grand challenge is to redesign evaluation so that it reflects clinical reality rather than linguistic convenience. Current translation and speech studies increasingly incorporate severity judgements, but multilingual healthcare still relies too heavily on metrics and protocols that are insufficient for deployment decisions. Ray et al. \shortcite{Ray2025} and Martos et al. \shortcite{Martos2025} illustrate why: a system can appear excellent under one evaluation setup and far more fragile under another, especially once multiple languages are included.

Risk-sensitive evaluation therefore requires at least four shifts. Errors must be weighted by potential clinical consequence. Outcomes must include comprehension, actionability, and correct follow-through rather than expert text comparison alone. Reporting must be stratified by language, dialect, accent, and task. And evaluation must become continuous rather than purely pre-deployment. In this sense, multilingual healthcare needs not just better benchmarks, but monitoring and auditing infrastructures aligned with real use \cite{mellingerChapter2Designing2025}. Are we doing this in MT?

\subsection{End-to-end multilingual fidelity}
The second challenge is end-to-end multilingual fidelity. Much of the literature still evaluates ASR, MT, simplification, summarisation, or note generation separately. In real healthcare workflows, however, these components increasingly operate in sequence. Spoken encounters may pass through recognition, translation, diarisation, summarisation, and note insertion, while written workflows may combine retrieval, translation, rewriting, and messaging. Each stage creates new opportunities for distortion.

The strongest implication of the spoken-language literature is that acceptable performance at one stage does not guarantee faithful end-to-end outcomes. The field therefore needs to move beyond component benchmarking and towards auditing how information changes as it passes through complete multilingual pipelines. And how errors stack after each step.

\subsection{Bounded agency and safe failure}
The third grand challenge is to ensure that flexible systems fail safely. Agentic systems are attractive because they can handle multiple subtasks, but that same flexibility makes them risky in multilingual care. A single system may translate, simplify, summarise, prioritise, and route information. In high-stakes settings, that is too much responsibility for an unbounded tool.

A strong research agenda on bounded agency would ask which tasks can be safely combined, where explicit hand-offs are required, how uncertainty should trigger abstention or escalation, and how intermediate outputs can remain visible for review, for example via quality estimation \cite{sindhujanReferenceLessEvaluationMachine2025}. In multilingual healthcare, safe failure should be treated as a first-class design objective rather than as an afterthought.

\subsection{Redesign of human roles and competencies}
The fourth challenge is the redesign of human roles and competencies. As AILTs enter healthcare, translators, interpreters, clinicians, informaticians, and administrative staff are not disappearing from multilingual communication workflows, but their roles are changing. The literature on PE workflows suggests that language professionals remain central, but increasingly as reviewers, terminology stewards, quality controllers, and escalation points \cite{briva-iglesiasLanguageEngineerTransversal2022,ehrensberger-dowNewRoleTranslators2023}. At the same time, clinicians are becoming more direct users of AI-mediated communication tools, often without systematic training \cite{marwahaAlgorithmicConsultantNew2025}.

This creates a dual need. People require practical AI literacy \cite{longWhatAILiteracy2020}: an understanding of likely failure modes, appropriate reliance, and the limits of automation. They also require stronger integration into workflow design and governance. Health systems, in turn, need explicit pathways specifying when professional interpreters, translators, or clinical reviewers must be involved. This challenge links translation studies, HCI, workforce redesign, and implementation science.

\subsection{Equity for minor languages, dialects, and accents}
The fifth challenge is equity. Across the literature, the strongest gains are least secure precisely where communication vulnerability is often greatest: minor languages, accent-diverse speech, and sociolinguistically marginalised communities. Current systems do not fail evenly, and this is not a minor technical inconvenience. It is a structural risk fulfilling the inverse care law, which states that the availability of good medical care tends to vary inversely with its need \cite{hartINVERSECARELAW1971}.

If healthcare organisations adopt AILTs because they work well for dominant languages, they may reduce cost and waiting times for some patients while entrenching lower-quality service for others. Equity therefore cannot be treated as a later optimisation problem. It must be built into evaluation, procurement, deployment, and reporting from the outset through stratified quality reporting, curated data strategies, community involvement, and explicit thresholds below which digital support should not displace professional services.

\subsection{Governance, regulation, and reporting}
The sixth challenge is governance. If multilingual communication is treated as clinical infrastructure, then failures in language mediation must become governable in ways analogous to other patient-safety issues. That requires institutional policies on approved use cases, audit trails, terminology management, version control, privacy safeguards, and incident reporting \cite{Shneiderman2020}. It also requires regulatory thinking capable of addressing systems whose behaviour varies with prompts, retrieval sources, context windows, and workflow integrations.

Broader AI-in-healthcare frameworks such as SPIRIT-AI and CONSORT-AI provide useful foundations for clearer reporting \cite{CruzRivera2020,Liu2020}, and Tan et al. \shortcite{Tan2026} push further by arguing that agentic systems require a different regulatory lens. What remains underdeveloped is a multilingual health communication perspective within these frameworks. The field still lacks strong norms for reporting language-specific variability, communication outcomes, and the boundary between linguistic and clinical responsibility.

\subsection{Trust-oriented UX and overreliance prevention}
The seventh challenge concerns trust and interface design. The literature repeatedly suggests that fluent output invites overreliance \cite{robinetteOvertrustRobotsEmergency2016}. What multilingual healthcare needs, however, is not maximal trust, but calibrated trust \cite{zerilliHowTransparencyModulates2022}. Users should be able to see what the system has done, what remains uncertain, and when review or escalation is necessary (see "seams" within HCI literature \cite{ehsanSeamfulXAIOperationalizing2024}).

For multilingual healthcare, this has direct interface implications. Systems should surface provenance when retrieval is involved, highlight unresolved terminology, flag uncertainty around clinically critical entities such as medication, dosage, negation, and time, and make editing and error reporting straightforward. Poor interface design can make even a technically strong model unsafe by encouraging autopilot behaviour. Conversely, well-designed interfaces can make imperfect systems safer by supporting reflection, verification, and contestation.

Taken together, these seven challenges suggest that the next phase of research should focus less on isolated tools and more on accountable multilingual communication infrastructures. The future of the field will depend not on whether AI can generate language, but on whether it can be embedded into systems that remain reliable, safe, and trustworthy under real-world conditions \cite{reiterWeShouldEvaluate2025}.

\section{Conclusions and research agenda}
Recent literature on AI language technologies in multilingual healthcare justifies both optimism and restraint. The optimism is real. For some written tasks and some language pairs, LLM-based translation is already approaching professional quality \cite{Ray2025}. Reviewed PE workflows can accelerate multilingual document production without abandoning expert oversight \cite{Brewster2025}. Plain-language rewriting can make difficult discharge materials more understandable \cite{Rust2025,Zaretsky2024}. Ambient documentation tools can reduce at least some dimensions of documentation burden and burnout \cite{Olson2025,Stults2025}. Domain-adapted speech pipelines can also improve performance in specialist settings \cite{IranzoSanchez2025}. These are meaningful advances for access, efficiency, and UX in multilingual healthcare.

The restraint is equally important. Gains are not distributed evenly across languages, accents, and settings. Spoken systems remain particularly fragile in real-world conversational conditions. Minor languages continue to face poorer performance. Agentic workflows advance faster rhetorically than empirically. And many of the hardest problems are institutional and not only technical: procurement, escalation pathways, role allocation, AI literacy, monitoring, and governance.

Consequently, the future of multilingual healthcare cannot be reduced to a competition between models and BLEU scores. It is a question of how communication systems are designed, how users are expected to rely on them, and how accountability is preserved when language mediation becomes increasingly automated. This is also why the paper belongs squarely within a Translators and Users perspective. Translators, interpreters, clinicians, and patients are not peripheral to these technologies. They are the actors through whom reliability, safety, and trustworthiness are either realised or undermined. AILT research in healthcare therefore needs to examine not only outputs, but also user roles, revised workflows, human oversight, and the conditions under which trust becomes warranted.

If HCAILT is used as a lens \cite{BrivaIglesias2026}, a practical research agenda for the next phase can be organised around three linked pillars. First, a reliability pillar should develop risk-sensitive multilingual benchmarks, end-to-end pipeline evaluations, and comprehension-oriented assessments grounded in real healthcare tasks. Second, a safety-culture pillar should focus on implementation: role design, AI literacy, escalation, terminology governance, and multilingual incident reporting. Third, a trust-and-governance pillar should develop interface patterns for calibrated use, reporting norms for multilingual AI interventions, and regulatory approaches that acknowledge the realities of non-deterministic and increasingly agentic systems.

The main implication of this review is therefore transdisciplinary. MT/NLP researchers contribute modelling and evaluation expertise. Translation studies contributes long-standing knowledge about mediation, revision, function, and language inequity. Healthcare researchers and practitioners ground the field in clinical reality. HCI contributes user-centred design and trust calibration. And policymakers shape the institutional conditions under which these systems can be used responsibly. No single field can solve these problems alone.

This review has limitations. It is a narrative synthesis rather than a formal systematic review, and it reflects a rapid-moving field, especially for agentic workflows, where discourse still outpaces strong deployment evidence. Even so, the present moment is decisive. Multilingual healthcare is one of the first domains in which AI language technologies are becoming infrastructure. The choices made now about evaluation, workflow design, governance, and equity will determine whether that infrastructure becomes merely efficient, or genuinely safe, just, and trustworthy.

\bibliographystyle{eamt26}
\bibliography{eamt26}
\end{document}